# An Enhanced Geometric-Spectral Feature Learning Framework for Airborne Multispectral Point Cloud Classification

Xian Li, *Member, IEEE*, Yanfeng Gu, *Senior Member, IEEE*, Linghua, Xu, Aleksandra Pižurica, *Senior Member, IEEE*

***Abstract*—Multispectral point cloud (MPC) is composed of 3D spatial-spectral information, which holds tremendous potential for accurate land-cover classification. However, the representation power of classification models is limited by inherent high-dimensional and heterogeneous spatial–spectral information, unbalanced sample distribution, and inter-class spectral similarity of airborne MPCs. We build two MPC datasets and propose an enhanced geometric-spectral feature learning framework based on attentions for airborne MPC classification. A key component in our model is a two-stream feature fusion method with attention mechanisms, which enhances the representation capability of spatial–spectral features from high-dimensional heterogeneous MPCs. The first stream aims to extract position-encoded global spectral features with fusion self-attention, and the second stream comprises a multikernel point convolution and feature aggregation attention to extract spectral-guided geometric features. We then develop a residual attention fusion block to integrate the most informative geometric-spectral features from the two parallel streams. Another important contribution of this work is a joint loss function to improve the learning ability on unbalanced and inter-class similar samples. Experimental results on two airborne MPC datasets demonstrate the effectiveness of the proposed method compared with the state-of-the-art methods. Furthermore, the codes and datasets used in this paper will be made available freely at https://github.com/HITlixian/TGRS_GSFF.**

## I. INTRODUCTION

Multispectral point cloud (MPC) is a novel type of point cloud data that each point vector encompasses three-dimensional (3D) positional attributes and several to dozens of spectral bands. Compared with LiDAR data or multispectral imagery, MPC is composed of both 3D spatial and spectral information [1], [2], which holds tremendous potential for a range of applications, including smart agriculture [3], [4], forest inventory [5], underwater identification [6], and military surveillance [7], [8].

MPC can be generated by four different modes: multi-wavelength LiDAR [9], multispectral LiDAR [10], [11], multispectral imaging [1], [2], multispectral and LiDAR [12], [14]. A renowned multi-wavelength LiDAR is the Teledyne Optech Titan system, which consists of three independent laser channels (532, 1064, and 1550 nm) at different angles. MPC can be generated by using multi-wavelength point cloud fusion methods [9], [16]. Multispectral LiDAR utilizes a supercontinuum laser or a tunable laser associated with waveform decomposition methods to generate MPC with a broadband spectrum [10], [11]. The primary advantage of Multispectral LiDAR is that it captures both the geometry and spectral signatures of the observed scenes simultaneously. Overlapped multispectral images collected by a multispectral sensor can be generated to MPC by using image-matching 3D reconstruction techniques [1], [2], [13]. Multispectral and LiDAR data are combined with an image-point cloud registration method to generate MPC [14], [15].

Next to MPC generation, various methods for MPC processing have been reported recently [17], [18], [19]. Land-cover classification of MPCs is one of the most fundamental and demanding tasks since it is an essential step in a range of applications. MPC data processing methods can be classified into three modes: projection-based [20], [21], [22], voxel-based [23], and point-based [9], [24], [25]. Projection-based mode typically rasterizes irregular MPC into structured data by projecting or quantizing, and then employs traditional multispectral image processing methods for classification. This inevitably results in spatial information loss in some categories and a significant computational burden [26]. An alternative approach is the voxel-based mode, which transforms irregular MPC to regular 3D voxelization representations [23], followed by 2D or 3D convolution operations. Similarly, this mode incurs information loss and computational burden due to the cubic growth in the number of voxels as a function of resolution. Recent approaches to improve the computational efficiency include sparse convolution [27] and the structured state space model [23].

Manuscript received XX XXX 2026; revised XX XXX 2026; accepted XX XXX, 2026. Date of publication XX XXX 2026; date of current version XX XXX 2026. This work was supported by National Natural Science Foundation of China through the Youth Science Foundation Project (Grant No. 62301192), Heilongjiang Science Foundation for Excellent Young Scholars (Grant No. YQ2024F003), and Postdoctoral Special Science Foundation of China (Grant No. 2024T171158).

Xian Li, Yanfeng Gu, and Linghua, Xu are with the School of Electronics and Information Engineering, Harbin Institute of Technology, Harbin 150001, China (e-mail: xianli@hit.edu.cn; guyf@hit.edu.cn).

Aleksandra Pižurica is with the Department of Telecommunications and Information Processing, UGent-GAIM, Ghent University, 9000 Ghent, Belgium.

Compared with projection-based or voxel-based modes, the point-based mode operates directly on irregular MPC with designed deep learning models in 3D continuous space, extracting spatial and spectral features simultaneously. These models can be grouped into three main categories: multilayer perceptrons (MLP) [28], [29], Graph convolutional networks (GCN) [9], [30], and point transformers [31], [32], [33]. The first pioneering deep learning model is PointNet [28], which exploits pointwise MLP and pooling layers with permutation-invariant characteristics to extract and aggregate local spatial and spectral features from MPC. Subsequently, it was extended to PointNet++ [29], PointConv [35], RandLA-Net [36], PointNeXt [37], et al. GCN aims to connect the point set into a graph and performs graph convolutions to construct local spectral and geometric relationships [9], [38]. Point transformer replaced the MLPs and GCN with a self-attention mechanism to learn long-range context dependencies. Recently, it was optimized into grouped self-attention [32], a faster version [33], and a lightweight version [34]. Lately, several Mamba networks [23], [39], [40] were proposed to improve global context modeling capability with linear computational complexity.

Although the above-described methods demonstrated huge success in MPC classification, two important challenges remain. The primary challenge is how to integrate 3D spatial–spectral information more effaectively. This is because 3D positional and spectral information have fundamentally different characteristics, while multispectral information inherently presents high-dimensional and nonlinear characteristics. Existing methods generally fuse such high-dimensional heterogeneous spatial and spectral information via simple feature concatenation [28], [29] or graph convolution operations [9], [38], which limits their feature representational capability and degrades MPC classification accuracy. The second challenge is how to address the problems of long-tail distribution and semantic ambiguity of airborne MPCs in real outdoor scenes observed by airborne platforms [24], [41], [42]. Current works in MPC classification, including [25], [30], [36], typically yield lower classification accuracy for minority and spectrally similar classes under the long-tailed distribution (i.e., unbalanced sample distribution between different classes) and semantic ambiguity (i.e., inter-class spectral similarity due to spatial variability of spectral features [43] ) cases, especially when the number of labeled training data is limited and sparse. Therefore, how to design an enhanced geometric-spectral feature learning framework that can adapt well to airborne MPCs is not evident.

Motivated by these challenges, we build two airborne MPC datasets and propose an enhanced geometric-spectral feature learning framework with attention for land-cover classification. Specifically, we devise a two-stream feature fusion method, which simultaneously extracts discriminative local geometric and global spectral features in parallel. The first stream aims to extract spatial position-aware global spectral features from MPC with a fusion self-attention, where the second stream captures geometric features from relative spatial position information under the guidance of the extracted global spectral features with a feature aggregation attention and multikernel point convolution. Thus, the proposed two streams interact with each other during the training process in an end-to-end manner from scratch, enhancing the geometric-spectral feature learning capability. Moreover, we develop a residual attention fusion method to further integrate discriminative geometric-spectral features from the two parallel streams. Then, we formulate a lightweight encoder-decoder classification framework based on a fully point convolution network, and each encoder layer applies the devised two-stream feature fusion method. Finally, compared with the common cross-entropy loss that treats all classes uniformly, we define a joint loss function to enhance intra-class compactness and inter-class separability, mitigating the problems of long-tail distribution and semantic ambiguity in airborne MPCs.

The main contributions of this article are given as follows:

1) We propose a lightweight geometric-spectral feature learning framework based on point convolution for airborne multispectral point cloud classification. The main advantage of the proposed framework is its powerful 3D spatial–spectral feature learning capability, exhibited as an improved performance compared with the current state-of-the-art.

2) We devise a position-encoded global spectral feature extractor and a spectral-guided geometric feature extractor in parallel, which are integrated with a residual attention to learn discriminative geometric-spectral features. This interactive configuration effectively learns geometric-spectral features.

3) We introduce an effective joint loss function that incorporates a long-tail center loss and a minimum margin loss to enforce intra-class compactness and inter-class separability. This joint loss can be applied to other networks too and is especially of interest when the number of labelled training data is limited and sparse.

4) We build two MPC datasets characterized by a long-tailed distribution and semantic ambiguity for land-cover classification, which will be openly available after a possible publication, contributing to the remote sensing community.

The rest of this article is organized as follows. Section II introduces the proposed method. Section III evaluates the effectiveness of the proposed method on real multispectral point cloud datasets, and Section IV draws the conclusion.

## II. METHODOLOGY

Let $\mathbf{M}=\{\mathbf{m}_n\}_{n=1}^{N} \in \mathbb{R}^{N\times(3+B)}$ denote an MPC sample with $N$ points, where each point $\mathbf{m}_n \in \mathbb{R}^{1\times(3+B)}=[\mathbf{p}_n,\mathbf{f}_n]$ contains heterogeneous 3D positional attributes $\mathbf{p}_n \in \mathbb{R}^{1\times 3}=(x,y,z)$ and multispectral information with $B$ spectral bands $\mathbf{f}_n \in \mathbb{R}^{1\times B}=[f_n^1,\cdots,f_n^B]$. Suppose that $T$ points in MPC are labeled and denote the labeled set by $\mathcal{T}=\{\mathbf{m}_n,r_n\}_{n=1}^{T}$, where $r_n$ denotes the class label from the set $\mathcal{C}=\{1,\cdots,C\}$, where $C$ is the number of classes. For a given MPC, our goal is to predict the labels of unlabeled points.

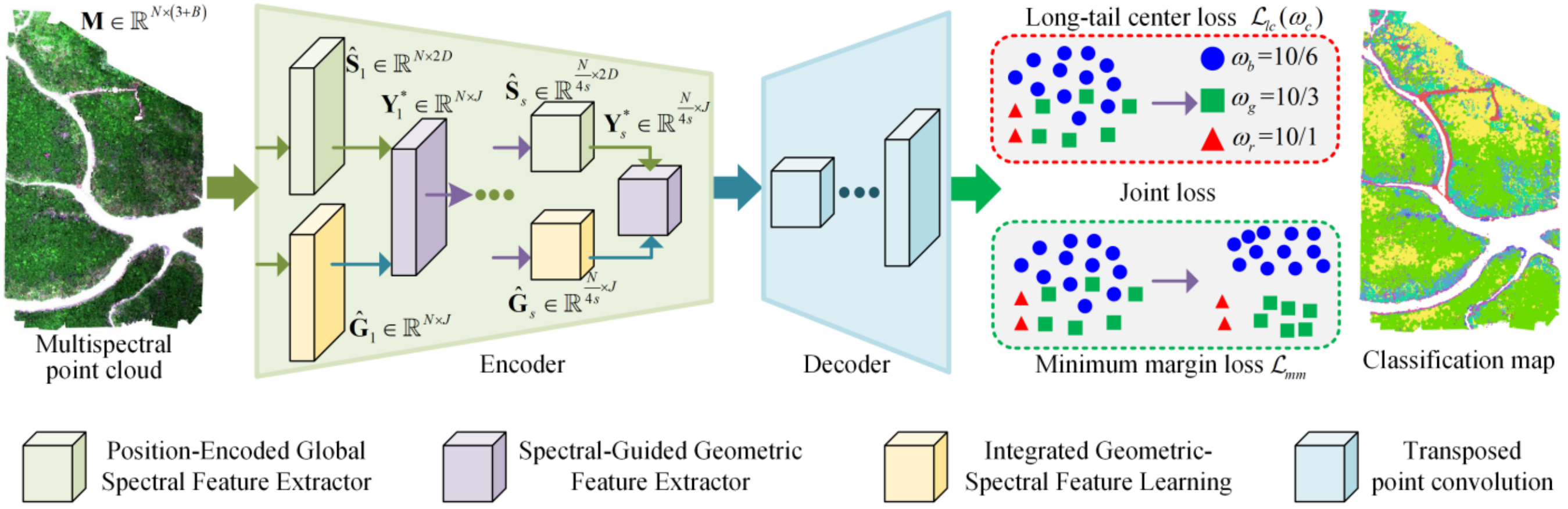

**Fig. 1.** Overall framework of the proposed method. It is a lightweight encoder-decoder network with a full point convolution, which is trained with the joint lossin an end-to-end fashion from scratch. Each encoder layer from stage 1 to $S$ is a two-stream feature fusion method that comprises three key components: position-encoded global spectral extractor, spectral guided geometric feature extractor, and integrated geometric-spectral feature learning.

### *A. Overview*

The primary challenge of MPC classification is extracting discriminative features from high-dimensional and heterogeneous 3D position-spectral information, which is exacerbated by long-tailed distribution and semantic ambiguity. We propose a geometric-spectral feature learning framework based on attention for airborne MPC classification, which enhances the capability of heterogeneous feature representation. In this framework, we formulate a lightweight encoder-decoder network with point convolution. Each encoder layer is a two-stream feature fusion method that contains three key components. The first component with a fusion self-attention is to extract position-encoded global spectral features from MPC, where the second component performs a spectral-guided geometric feature extraction powered by a feature aggregation attention in parallel. The third component integrates the two parallel components with a residual attention to learn discriminative geometric-spectral features. A joint loss that incorporates long-tail center loss and minimum margin loss is devised to address the problems of long-tail distribution and semantic ambiguity. Fig. 1 shows the overall framework of the proposed method. In the following, we present the details of the three components and the joint loss, as shown in Fig. 2.

### *B. Position-Encoded Global Spectral Feature Extractor*

Due to unstructured spatial distribution and heterogeneous nature of 3D positional and spectral information in MPC, inspired by the success of point transformer [31], [33], [39], we here devise a position-encoded spectral feature extractor to fully fuse local neighbor spatial and global spectral correlation features, facilitating heterogeneous spatial-spectral feature aggregation. The main idea is to encode the MPC samples into their relative spatial position and global spectral correlation representations in parallel using the 2D point convolutions, which ensures feature dimensionality matching while preserving the spatial geometric structure of MPC, and then to feed a self-attention mechanism and max pooling, achieving spatial position-aware global spectral feature learning.

The input is a matrix $\mathbf{M}\in\mathbb{R}^{N\times(3+B)}=[\mathbf{P},\mathbf{F}]$, obtained by directly stacking an unordered MPC sample. Considering the strong spectral-spatial correlation among spatially adjacent points, $(K-1)$ local neighborhood points for each point in $\mathbf{M}$ are selected with the K-Nearest Neighbor (KNN) method for efficiency, denoted by $\{\mathcal{P}\in\mathbb{R}^{N\times K\times 3},\mathcal{F}\in\mathbb{R}^{N\times K\times B}\}$.

*1) Global Spectral Feature Extraction*: The global spectral feature extraction block in our architecture exploits a 2D point convolution layer to capture global spectral correlation features and preserve the spatial geometric structure of MPC. While earlier reported spectral feature extraction methods were always based on MLP [24], [33], [44], our method captures global spectral features via a lightweight 2D point convolution with shared weights. The global spectral feature matrix in the $d$th feature map $\hat{\mathbf{F}}_d\in\mathbb{R}^{N\times K}$ of the first block is computed by:

$$\hat{\mathbf{F}}_d=\delta(\mathcal{W}_d*\mathcal{F})=\delta\left(\sum_{b=1}^{B}w_d^b\cdot\mathbf{F}^b\right) \tag{1}$$

where $\delta(\cdot)$ denotes the leaky ReLU activation function, $\mathcal{W}_d\in\mathbb{R}^{1\times1\times B}=[w_d^1,w_d^2,\ldots,w_d^B]$ is the 2D point convolutional kernel connected to the $d$th feature map, * is the convolution operation. $\mathbf{F}^b\in\mathbb{R}^{N\times K}$ is the $b$th spectral map of $\mathcal{F}$. The bias terms are omitted in this manuscript. Equation (1) indicates that each global spectral feature is extracted from the whole spectral information $\mathcal{F}=[\mathbf{F}^1,\cdots,\mathbf{F}^b,\cdots,\mathbf{F}^B]$ with the shared weights $\mathcal{W}_d$, which requires fewer parameters compared to traditional methods with MLP. We utilize multiple kernels to extract different global spectral features. The output of the first block is $\hat{\mathcal{F}}\in\mathbb{R}^{N\times K\times D}=[\hat{\mathbf{F}}_1,\cdots,\hat{\mathbf{F}}_d,\cdots,\hat{\mathbf{F}}_D]$, where $D$ is the number of kernels.

*2) Local Spatial Position-Encoding*: To adapt to the unstructured spatial distribution in MPC, we introduce a self-learned local spatial position encoding block using a 2D point convolution layer. The use of this block enables maintaining the spatial geometric structure and position awareness during the whole training process. We define the local spatial position-encoding matrix in the $d$th feature map $\hat{\mathbf{P}}_d \in \mathbb{R}^{N\times K}$ of the first block as follows:

$$\hat{\mathbf{P}}_d = \delta(\mathcal{W}_d^p * (\bar{\mathcal{P}} \oplus (\mathcal{P} - \bar{\mathcal{P}}) \oplus d(\mathcal{P}, \bar{\mathcal{P}}))) \quad (2)$$

where $\bar{\mathcal{P}} \in \mathbb{R}^{N\times K\times 3}$ is the repetition of $\mathbf{P}$ for size matching with $\mathcal{P}$. $\oplus$ and $d(\cdot,\cdot)$ denote the operations of concatenating and Euclidean distance calculation that maintains the original dimensionality, respectively. $\mathcal{W}_d^p \in \mathbb{R}^{1\times 1\times 7}$ is the corresponding 2D point convolutional kernel with shared weights, which encodes $(\bar{\mathcal{P}} \oplus (\mathcal{P} - \bar{\mathcal{P}}) \oplus d(\mathcal{P}, \bar{\mathcal{P}}))$ in a pointwise way to preserve the spatial geometric structure of MPC. Similarly, we use multiple kernels to establish different position-encoding features $\hat{\mathcal{P}} \in \mathbb{R}^{N\times K\times D} = [\hat{\mathbf{P}}_1, \hat{\mathbf{P}}_2, \cdots\cdots, \hat{\mathbf{P}}_D]$.

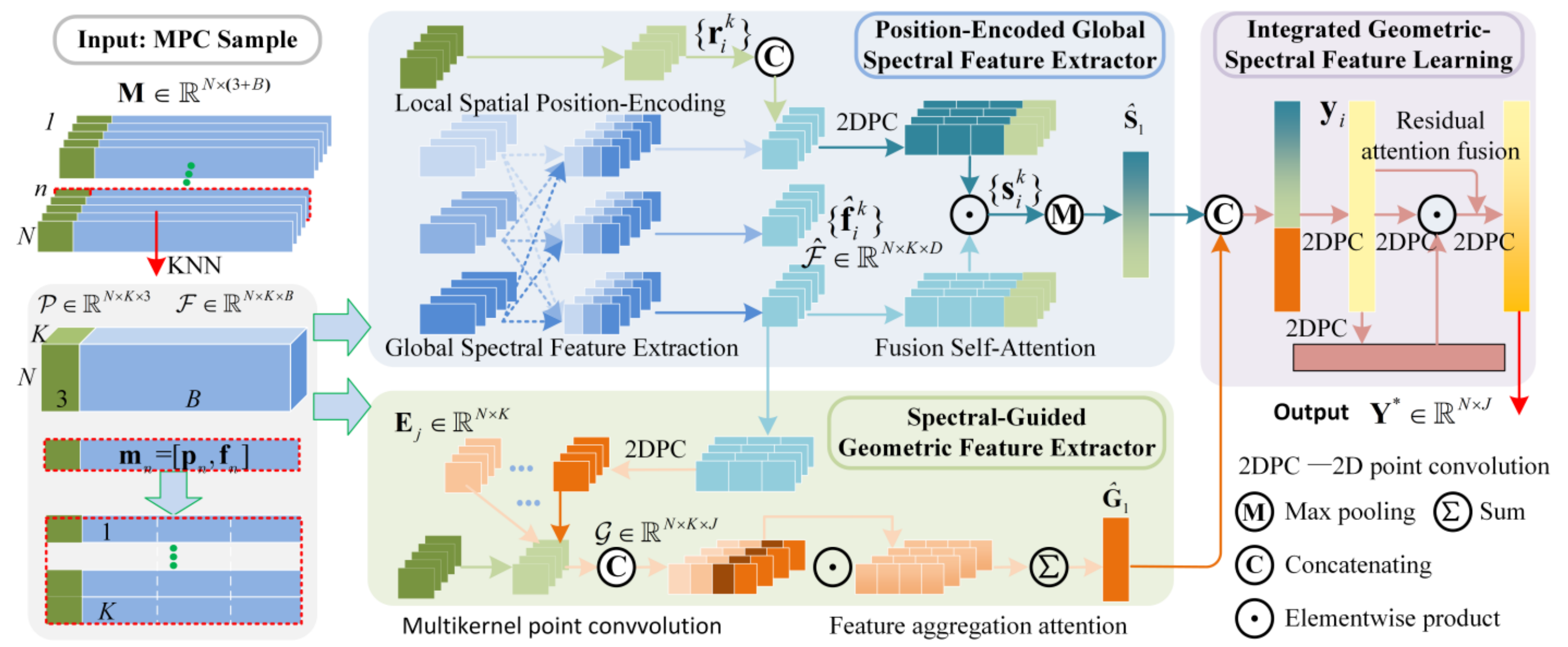

**Fig. 2.** Illustration of the proposed position-encoded global spectral extractor, spectral guided geometric feature extractor, and integrated geometric-spectral feature learning.

*3) Spectral-Position Fusion Self-Attention*: To fully fuse the extracted global spectral and position-encoding features, we devise a spectral-position fusion self-attention mechanism based on 2D point convolution, which encodes the position information into the extracted global spectral features, learning the spatial position-encoded global spectral feature matrix $\mathbf{S}_d \in \mathbb{R}^{N\times K}$:

$$\mathbf{S}_d = \sigma(\mathcal{W}_d^f * ((\hat{\mathcal{F}} - \bar{\mathcal{F}}) \oplus \hat{\mathcal{P}})) \odot (\hat{\mathcal{F}} \oplus \hat{\mathcal{P}}) \quad (3)$$

where $\mathcal{W}_d^f \in \mathbb{R}^{1\times 1\times 2D}$ denotes the lightweight 2D point convolutional kernel with shared weights. $\bar{\mathcal{F}} \in \mathbb{R}^{N\times K\times D}$ is the repetition of the 1st feature map $\bar{\mathbf{F}}_1 \in \mathbb{R}^{N\times D} = \hat{\mathcal{F}}_{k=1}$. $\sigma(\cdot)$ is the sigmoid activation function for normalization. $\odot$ denotes the elementwise product. Equation (3) shows that the concatenating position-encoding and global spectral features are recalibrated with a self-learned attention matrix, which is established from position-embedded $\hat{\mathcal{P}}$ and spectral feature differences $(\hat{\mathcal{F}} - \bar{\mathcal{F}})$ with $\mathcal{W}_d^f$. Similarly, we use multiple kernels to establish different position-encoding global spectral features. The output of the first spectral-position fusion self-attention is $\mathcal{S} \in \mathbb{R}^{N\times K\times 2D}$. By introducing the elementwise maximum operation, the output of the position-encoded global spectral feature extractor in the stage 1 is defined as $\hat{\mathbf{S}}_1 \in \mathbb{R}^{N\times J}$.

## *C. Spectral-Guided Geometric Feature Extractor*

To further enhance the representation capability of heterogeneous positional and spectral information in MPC, we design a spectral-guided geometric feature extractor based on the spectral similarity principle, to capture geometric correlation features from each center point and its local neighborhood points. The main idea is to perform 2D multikernel point convolution to encode spectral feature differences into a set of spectral guidance tensors that fuse with the corresponding relative spatial position, and then to feed a feature aggregation attention to output informative geometric features. In this way, we aim to formulate the geometric correlation features under the guidance of the extracted global spectral features, thereby boosting the geometric-spectral feature learning capability.

To establish the spectral guidance tensors, we embed the spectral feature differences from the global spectral feature extraction block. This is achieved by using a 2D point convolutional layer. We define the spectral guidance matrix in the $j$th kernel $\mathbf{E}_j \in \mathbb{R}^{N\times K}$ as follows:

$$\mathbf{E}_j = \sigma(\mathcal{W}_j^s * (\hat{\mathcal{F}} - \bar{\mathcal{F}})) \quad (4)$$

where $\mathcal{W}_j^s \in \mathbb{R}^{1\times 1\times B}$ is the $j$th kernel of the 2D point convolutional layer with shared weights. We use $J$ kernels to generate the spectral guidance tensor. Subsequently, we perform triplekernel point convolution to establish triple independent spectral guidance tensors $\mathcal{E} \in \mathbb{R}^{N\times K\times 3\times J}$, which are fused with the relative spatial position to measure the geometric correlation $\mathcal{G} \in \mathbb{R}^{N\times K\times J}$ as follows:

$$\mathcal{G} = \delta\left\langle \mathcal{E}, (\mathcal{P} - \bar{\mathcal{P}}) \right\rangle \quad (5)$$

where $\langle \mathcal{A}, \mathcal{B} \rangle$ denotes batched matrix product of $\mathcal{A}$ and $\mathcal{B}$.

To emphasize informative geometric correlation features and suppress less useful ones, we design a feature aggregation attention as follows:

$$\hat{\mathbf{G}} = \sum_{k=1}^{K} \mathcal{G}_k \odot \sigma\left(\mathcal{W}^g * \mathcal{G}\right)_k \quad (6)$$

where $\mathcal{W}^g$ denote the kernels of 2D point convolution for the feature aggregation attention.

### *D. Integrated Geometric-Spectral Feature Learning*

Having extracted position-encoded global spectral features and spectral-guided geometric features from the two dedicated feature extractors, we introduce a residual attention fusion method to effectively integrate them. Specifically, we employ a 1D point convolution to fuse the extracted geometric-spectral features, and then introduce a residual attention mechanism to further enhance the feature representation capability. The fused geometric-spectral feature vector $\mathbf{y}_j \in \mathbb{R}^{N\times 1}$ is computed as

$$\mathbf{y}_j = \delta(\mathbf{W}_j * (\hat{\mathbf{S}} \oplus \hat{\mathbf{G}})) \quad (7)$$

where $\mathbf{W}_j \in \mathbb{R}^{1\times(2D+J)}$ is the kernel matrix of 1D point convolution. Equation (7) indicates that each fused feature $\mathbf{y}_j$ is extracted from all the geometric-spectral features in $(\hat{\mathbf{S}} \| \hat{\mathbf{G}})$ with $\mathbf{W}_j$ and $\delta(\cdot)$, to form integrated geometric-spectral features. The output can be expressed by: $\mathbf{Y} \in \mathbb{R}^{N\times J}$.

We introduce a residual attention mechanism that exploits a 1D point convolutional layer with sigmoid activation function to transform $\mathbf{Y}$ into a normalized importance weight, which is used to adaptively emphasize informative features of $\mathbf{Y}$, and then employs a residual learning strategy to facilitate network training and generate discriminative geometric-spectral features as follows:

$$\mathbf{Y}_1^* = \sigma\left(\mathbf{W} * \mathbf{Y}\right) \odot \mathbf{Y} + \mathbf{Y} \quad (8)$$

where $\mathbf{W}$ is the kernel matrix of the 1D point convolution.

By the proposed method, the heterogeneous data comprising 3D position and spectral information of the input $\mathbf{M} \in \mathbb{R}^{N\times(3+B)}$ is elegantly encoded into feature representations $\hat{\mathbf{S}}_1 \in \mathbb{R}^{N\times 2D}$ and $\hat{\mathbf{G}}_1 \in \mathbb{R}^{N\times J}$ via a position-encoded global spectral feature extractor and a spectral-guided geometric feature extractor, respectively, and are then integrated into the discriminative geometric-spectral features $\mathbf{Y}_1^* \in \mathbb{R}^{N\times J}$ using our residual attention fusion method, explicitly

### *E. Network Architecture and Joint Loss*

*1) Network Architecture:* Having developed the two-stream feature fusion method with three key components, we formulate a lightweight encoder-decoder network based entirely on point convolution, with four stages each, as shown in Fig. 1. The encoder has four stages that operate on progressively downsampled MPC with downsampling rate of 4, and each stage is composed of the three key components. The point-wise geometric-spectral features are restored using a point convolutional layer in each decoder stage. For classification, the final decoder stage output a feature vector for each point in MPC, we apply several point convolutional layers and a softmax function to map this feature to predict the outputs of labels.

*2) Joint Loss:* Now, we turn to define the joint loss function explicitly. Current works in MPC classification, including [9], [23], [33], often use the common cross-entropy loss, which treat all classes uniformly. However, airborne MPC often exhibits long-tail distribution and inter-class semantic ambiguity due to spatial variability of spectral features, a nonuniform loss function is more appropriate. Inspired by center loss [45], we here devise a long-tail center loss scheme that introduces a modulating factor on traditional center loss, which minimizes the distance of each point in a class to its center, enforcing intra-class compactness for all classes to address inherent intra-class spectral variability. We define the long-tail center loss $\mathcal{L}_{lc}$ as follows:

$$\mathcal{L}_{lc} = \sum_{c=1}^{C} \omega_c \sum_{n=1}^{N_c} \left\| \mathbf{z}_n^c - \mathbf{m}_c \right\|_2^2 \quad (9)$$

where $N_c$ is the number of training samples for the *c*th class, and $\omega_c$ is the modulating factor that is computed by $\left(\sum_{c=1}^{C} N_c\right) / N_c$, which enhances the regularization of the minority classes to mitigate long-tail distribution problem. $\mathbf{z}_i^c$ and $\mathbf{m}_c$ are the network output of the *n*th point and its centers, respectively.

To enforce inter-class separability, maximizing the embedding space distances among class centers is a nature scheme. We instead introduce a minimum margin loss that exploits the reciprocal and minimum function to increase the distance between two nearest centers. This is to prevent the proposed model from overfitting to the two largest centers. We define the minimum margin loss $\mathcal{L}_{mm}$ as follows:

$$\mathcal{L}_{mm} = \min_{i,j\in[1,C], i\neq j} \left\| \mathbf{m}_i - \mathbf{m}_j \right\|_2^2 \quad (10)$$

Inspired by logarithmic absolute error [46], we develop a joint loss $\mathcal{L}_{joint}$ as follows:

$$\mathcal{L}_{joint} = \log\left(1 + \mathcal{L}_{lc} / \mathcal{L}_{mm}\right) \quad (11)$$

In contrast to traditional center loss [45], which equally contributes to different classes without distance constraints among different class centers, leading to inefficient learning capability for minority and similar classes, we develop a joint loss that incorporates the long-tail center loss and minimum margin loss to improve intra-class compactness and inter-class separation, mitigating the problems of long-tail distribution and semantic ambiguity. The final loss is the weight sum of the joint loss and cross-entropy loss.

$$\mathcal{L} = \mathcal{L}_{CE} + \lambda \mathcal{L}_{joint} \quad (12)$$

where $\mathcal{L}_{CE}$ cross-entropy loss. $\lambda$ is the regularization parameter to balance the two components.

## III. Experimental Results and Analysis

We perform experiments on two airborne MPC datasets: Zhangjiangkou Mangrove (ZJKM) and Shankou Mangrove (SKM). Four objective metrics, overall accuracy (OA), average accuracy (AA), Kappa coefficient, and mean intersection over union (mIoU), are used for evaluation. For each experiment, we report the mean and standard deviation of the classification results over three runs with randomly selected training samples.

### A. Data Description and Hyperparameter Setting

The ZJKM dataset was collected on November 15, 2022, at the Zhangjiangkou Mangrove National Nature Reserve, Fujian Province, China. It comprises 771 multispectral images acquired by our asymmetric dual-angle multispectral stereo imaging system [1], with 60% image overlap along the flight path and 80% in the heading direction. The reconstructed ZJKM MPC contains 25,022,760 points, covering an area of 689 × 647 meters, featuring $B$=10 spectral bands spanning 430–870 nm at a spatial resolution of 5.5 cm. Seven land-cover classes were manually annotated: tidal flats (17.96%), buildings (0.14%), B. gymnorhiza (0.14%), A. corniculatum (0.21%), timber pavement (0.08%), A. marina (0.25%), and K. obovate (0.24%); The remaining 81.01% of points are unlabeled, resulting in sparse annotation due to the difficulty of human access. Among the seven classes, tidal flats dominate with 17.96%, while the other six classes are rare tail classes ranging from 0.08% to 0.25%. Notably, the smallest class (timber pavement) accounts for only about 1/224 of the majority class. Furthermore, the four mangrove species exhibit inter-class spectral similarity, attributed to both the spatial variability of spectral features and their shared taxonomic group. Consequently, this dataset suffers from a long-tail distribution and inter-class semantic ambiguity.

The SKM dataset was captured on October 19, 2021, at the Shankou Mangrove National Nature Reserve, Guangxi Zhuang Autonomous Region, China. It consists of 658 multispectral images captured by a multispectral sensor [2], with 60% overlap along the flight path and 80% in the heading direction. The reconstructed SKM MPC contains 35,058,937 points over a 427 × 567 meter area and shares the same spectral configuration as ZJKM with a spatial resolution of 4.0 cm. Eight land-cover classes were manually annotated: bridge (1.03%), B. gymnorhiza (0.55%), R. stylosa (0.29%), A. marina (0.15%), K. obovata (0.10%), bare earth (0.07%), A. corniculatum (0.05%), and boats (0.03%). With 97.73% of points unlabeled, the annotation sparsity is even more extreme than that of ZJKM. This dataset also exhibits a long-tail distribution (the largest class accounts for 1.03%, and the remaining six tail classes range from 0.03% to 0.55%) and inter-class semantic ambiguity among five mangrove species.

It should be noted that both MPC datasets are generated by using image-matching 3D reconstruction techniques [1], [2] as shown in Fig. 3. Their geometric information is reconstructed indirectly from multispectral images and may be affected by image-matching errors, occlusion, and local reconstruction uncertainty, compared with geometry-dominant LiDAR point clouds. Furthermore, the multispectral images collected in real outdoor scenes are inevitably influenced by sensor noise, illumination variation, and shadows, which may introduce data degradation and spectral variability [43]. The ground truths of the two datasets are shown in Fig. 4 (a) and Fig. 5 (a), respectively.

A grid-balanced sampling strategy [24] was applied to create the training, validation, and testing sets, which are separated spatially to avoid the point sharing problem. The receptive field was set to 4096. The number of training epochs and batch size were set to 300 and 8, respectively. The initial learning rate was empirically set to 0.01, which was multiplied by 0.95 at the end of every five epochs.

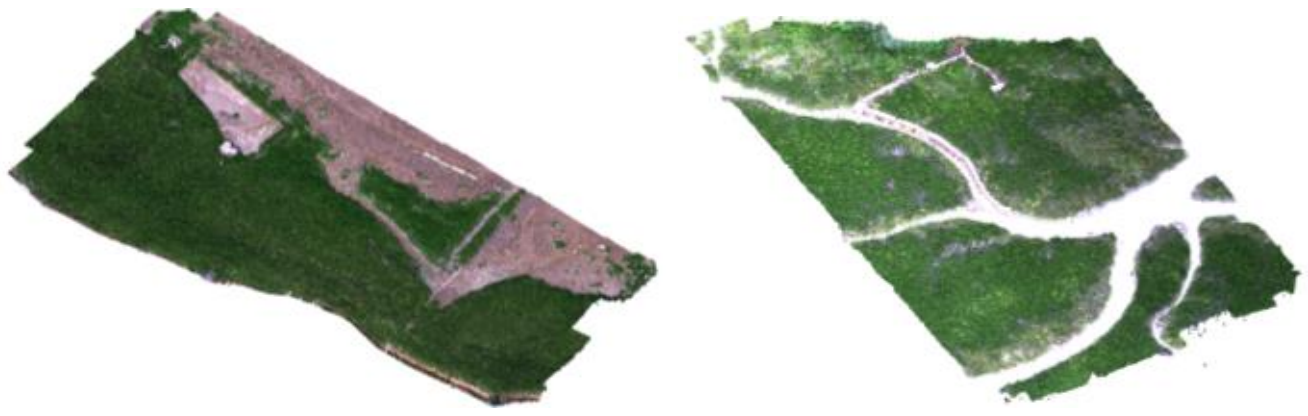
(a) (b)

**Fig. 3.** Two MPC datasets. (a) ZJKM. (b) SKM.

### B. Evaluation Results and Analysis

We compare the performance of the proposed model with the following representative methods, which are divided into three groups.

1) *LiDAR-Based Methods with Multispectral Input*: RandLA-Net [36], CGA-Net [48], RFFS-Net [30], EyeNet [49], MAMPC [25], PCMamba [40], PTv3 [33], and LitePT-S [34] are originally designed for LiDAR point clouds with accurate geometric and color attributes. To ensure a fair comparison, these methods simultaneously incorporate both geometric and spectral features from the two MPC datasets.
2) *Multispectral Imaging-Based Methods*: EMPC [24] is designed for multispectral point clouds generated from image matching techniques.
3) *Class Imbalance-Based Methods*: RandLA-Net [36] employs a weighted cross-entropy loss, RFFS-Net [30] uses a multi-level receptive field aggregation loss targeting small objects, PTv3 [33] and LitePT-S [34] adopt Lovász-Softmax loss to indirectly alleviate class imbalance by optimizing IoU, and EMPC [24] utilizes an adaptive loss incorporating a long-tail term.

For a fair comparison, the reference methods adopt the same sample configuration. Their hyperparameters are initialized according to their original implementations and, when necessary, fine-tuned on the validation set to ensure stable convergence on the two MPC datasets. Tables I and II report the class-specific accuracy, OA, AA, Kappa, and mIoU of the tested methods on the ZJKM and SKM datasets. The best result is marked in bold, and the second-best result is underlined.

As can be observed, the proposed method consistently yields the best OA, AA, kappa, and mIoU with a favorable improvement over the reference methods for the two datasets. For ZJKM (Table I), the proposed method yields OA of 87.82%, with the gains of 4.23%, 5.46%, 1.74%, 3.13%, 7.14%, 3.65%, 0.97%, 5.30%, and 1.62% over RandLA-Net, CGA-Net, RFFS-Net, EyeNet, MAMPC, PCMamba, PTv3, EMPC, and LitePT-S methods, respectively. The gains in OA, AA, kappa, and mIoU compared to the baseline methods are approximately 0.97% (over PTv3), 1.75% (over PTv3), 4.16% (over RFFS-Net), and 4.50% (over LitePT-S), respectively.

TABLE I
COMPARISON OF THE CLASSIFICATION ACCURACIES AMONG THE PROPOSED METHOD AND THE BASELINES USING THE ZJKM DATASET. THE BOLD ITEMS REPRESENT THE BEST VALUE, AND THE UNDERLINED ITEMS REPRESENT THE SECOND BEST VALUE

| Model | RandLA-Net (2020) | CGA-Net (2021) | RFFS-Net (2022) | EyeNet (2023) | MAMPC (2023) | PCMamba (2024) | PTv3 (2024) | EMPC (2025) | LitePT-S (2026) | **Ours** |
|---|---|---|---|---|---|---|---|---|---|---|
| Tidal flats | 99.18±0.02 | 98.94±0.24 | 99.89±0.04 | 99.60±0.11 | 93.99±3.45 | 98.03±2.57 | 98.98±0.13 | 95.39±2.75 | **99.90±0.04** | 97.69±0.77 |
| Building | 98.88±1.21 | 87.04±4.42 | 89.29±5.75 | 87.89±2.85 | 85.88±7.70 | 84.89±7.00 | 96.09±1.87 | 77.16±27.78 | **96.32±2.51** | 90.40±1.31 |
| B. gymnorhiza | 16.28±15.10 | 48.95±15.11 | 57.79±9.59 | 51.28±23.12 | **74.18±21.14** | 68.82±22.80 | 66.13±21.56 | 42.93±11.35 | 39.03 ± 8.22 | 69.71±7.83 |
| A. corniculatum | 33.59±9.45 | 49.17±3.51 | 64.25±2.12 | 62.01±9.17 | 65.68±7.48 | 70.41±3.49 | **89.75±1.42** | 47.34±4.29 | 52.17±25.04 | 73.96±3.03 |
| Timber pavement | 88.21±0.70 | 88.21±0.76 | 89.61±3.42 | 75.06±10.80 | 95.81±2.40 | 88.78±10.27 | 3.15±2.99 | **97.02±0.56** | 59.90±4.33 | 94.58±1.84 |
| A. marina | 80.32±10.01 | 62.95±7.65 | 66.02±7.29 | 76.73±3.68 | 74.96±10.54 | 66.26±14.26 | **89.80±1.08** | 80.35±16.90 | 67.33±13.82 | 82.92±6.28 |
| K. obovata | 80.99±3.71 | 81.47±1.57 | 65.13±2.09 | 74.79±5.39 | 44.82±20.45 | 80.13±22.53 | **87.23±2.41** | 82.17±9.09 | 74.93±2.07 | 72.03±6.26 |
| **OA** | 83.59±0.91 | 82.36±1.25 | 86.08±1.64 | 84.69±1.74 | 80.68±1.53 | 84.17±2.01 | 86.85±1.76 | 82.52±1.94 | 86.20±2.89 | **87.82±1.51** |
| AA | 71.06±2.53 | 73.82±1.65 | 76.00±2.53 | 75.34±4.68 | 76.47±3.34 | 79.62±4.14 | 81.29±2.61 | 74.62±5.78 | 73.52±3.29 | **83.04±2.42** |
| kappa | 77.79±1.40 | 76.21±1.72 | 79.44±2.47 | 79.09±2.36 | 73.98±2.17 | 78.23±2.50 | 55.70±1.95 | 75.96±3.17 | 75.14±4.82 | **83.60±2.03** |
| mIoU | 60.62±1.81 | 62.61±1.84 | 65.15±2.30 | 65.34±4.55 | 58.69±1.78 | 62.06±3.40 | 50.33±2.47 | 60.44±3.91 | 66.65±4.51 | **71.15±3.05** |

TABLE II
COMPARISON OF THE CLASSIFICATION ACCURACIES AMONG THE PROPOSED METHOD AND THE BASELINES USING THE SKM DATASET. THE BOLD ITEMS REPRESENT THE BEST VALUE, AND THE UNDERLINED ITEMS REPRESENT THE SECOND BEST VALUE

| Model | RandLA-Net (2020) | CGA-Net (2021) | RFFS-Net (2022) | EyeNet (2023) | MAMPC (2023) | PCMamba (2024) | PTv3 (2024) | EMPC (2025) | LitePT-S (2026) | **Ours** |
|---|---|---|---|---|---|---|---|---|---|---|
| Bridge | 99.23±0.15 | 98.95±0.14 | 99.20±0.36 | **99.79±0.01** | 87.89±3.96 | 94.51±0.89 | 98.38±0.21 | 96.96±0.62 | 85.09±2.06 | 97.88±0.32 |
| B. gymnorhiza | 89.23±4.16 | 84.66±0.56 | 89.28±2.54 | 90.82±0.96 | 79.13±5.22 | 86.58±0.52 | **94.18±1.33** | 90.57±0.83 | 92.63±0.82 | 85.93±3.85 |
| R. stylosa | 69.16±4.00 | 76.20±3.89 | 42.05±2.68 | 82.09±0.88 | 70.40±6.61 | 56.23±2.48 | 77.69±2.81 | 68.13±11.91 | **94.42±5.03** | 70.23±8.98 |
| A. marina | 51.19±16.90 | 58.44±9.63 | **76.22±7.60** | 50.69±2.25 | 60.33±1.78 | 60.77±3.65 | 64.60±1.57 | 66.04±4.98 | 26.59±5.39 | 75.74±2.60 |
| K. obovata | 24.37±13.15 | 19.12±6.57 | 50.01±12.04 | 11.58±9.70 | 39.22±5.75 | 11.93±4.29 | 47.97±2.92 | 22.14±9.19 | 31.79±9.89 | **55.00±6.17** |
| Bare earth | 86.00±4.04 | 91.35±2.93 | 87.22±5.17 | 87.90±2.14 | 96.08±1.25 | 77.85±12.87 | 41.06±1.09 | 87.39±8.52 | **99.80±0.01** | 88.64±2.01 |
| A. corniculatum | 24.07±10.98 | 12.76±8.67 | 33.54±10.45 | 21.97±2.06 | 59.26±9.61 | 21.44±6.66 | 29.65±2.38 | 49.08±10.03 | **68.62±12.31** | 55.15±9.73 |
| Boats | 95.13±3.99 | 95.14±5.72 | 42.24±6.54 | 96.37±1.16 | 99.67±0.45 | 98.67±0.69 | 86.00±1.74 | 99.94±0.09 | **100.00±0.00** | 99.62±0.32 |
| **OA** | 85.56±1.72 | 85.03±0.59 | 82.19±1.34 | 86.32±1.04 | 79.29±3.76 | 79.39±0.50 | 86.23±1.41 | 84.90±1.71 | 80.76±1.48 | **87.38±0.33** |
| AA | 67.30±2.74 | 67.08±1.55 | 64.97±2.65 | 67.65±2.11 | 74.00±1.47 | 63.50±1.49 | 67.44±2.65 | 72.53±3.12 | 74.87±1.00 | **78.52±1.25** |
| kappa | 78.81±2.54 | 78.21±0.77 | 75.25±1.82 | 80.45±1.40 | 72.06±4.60 | 70.88±0.59 | 81.18±1.86 | 78.73±2.53 | 66.61±3.02 | **82.22±0.52** |
| mIoU | 59.33±2.75 | 58.06±1.30 | 48.21±2.76 | 59.95±2.59 | 51.90±3.75 | 49.94±0.95 | 58.97±2.53 | 61.60±2.34 | 55.83±1.07 | **64.90±1.05** |

For SKM (Table II), the proposed method yields OA of 87.38%, with the gains of 1.82%, 2.35%, 5.19%, 1.06%, 8.09%, 7.99%, 1.15%, 2.48%, and 6.62% over RandLA-Net, CGA-Net, RFFS-Net, EyeNet, MAMPC, PCMamba, PTv3, EMPC, and LitePT-S methods, respectively. The gains in OA, AA, kappa, and mIoU compared to the baseline methods are approximately 1.06% (over EyeNet), 3.65% (over LitePT-S), 1.04% (over PTv3), 3.30% (over EMPC), respectively. This indicates that the proposed method has powerful geometric–spectral feature learning capability.

In terms of the class-specific accuracy, the proposed method performs the best accuracy only in one classe, but it yields comparable results to the best one in most of the classes for the two datasets. This is because the proposed joint loss incorporates a long-tail center loss to enhance the learning capability on the minority classes, sacrificing the accuracy on the majority classes to some extent. It is worth highlighting that the class-specific accuracy differs up to 82.90%, 49.99%, 42.10%, 48.32%, 50.99%, 31.77%, 95.83%, 54.09%, and 60.87% over RandLA-Net, CGA-Net, RFFS-Net, EyeNet, MAMPC, PCMamba, PTv3, EMPC, and LitePT-S methods from one class to another for ZJKM, respectively. For our method, this variation is only 27.98%. Similarly, for SKM in Table II, the class specific accuracy differs up to 75.16%, 86.19%, 65.66%, 88.21%, 60.45%, 86.74%, 68.73%, 77.80%, and 73.41% over RandLA-Net, CGA-Net, RFFS-Net, EyeNet, MAMPC, PCMamba, PTv3, EMPC, and LitePT-S methods from one class to another, respectively. For our method, this variation is only 44.47%. This indicates the superior robustness of the proposed method to different classes of the ZJKM and SKM datasets, both of which exhibit long-tail distributions and inter-class semantic ambiguity, which is an important asset for its practical applicability. Apart from quantitative analysis, Figs. 4-5 show the full classification maps, obtained by different methods on the two d atasets, and the detailed maps at one or two spatial positions (marked by red and gray boxes) are shown in the bottom. Visually, they are consistent with results the results reported in Tables I-II. Furthermore, the proposed method presents more similar results to the ground truth maps.

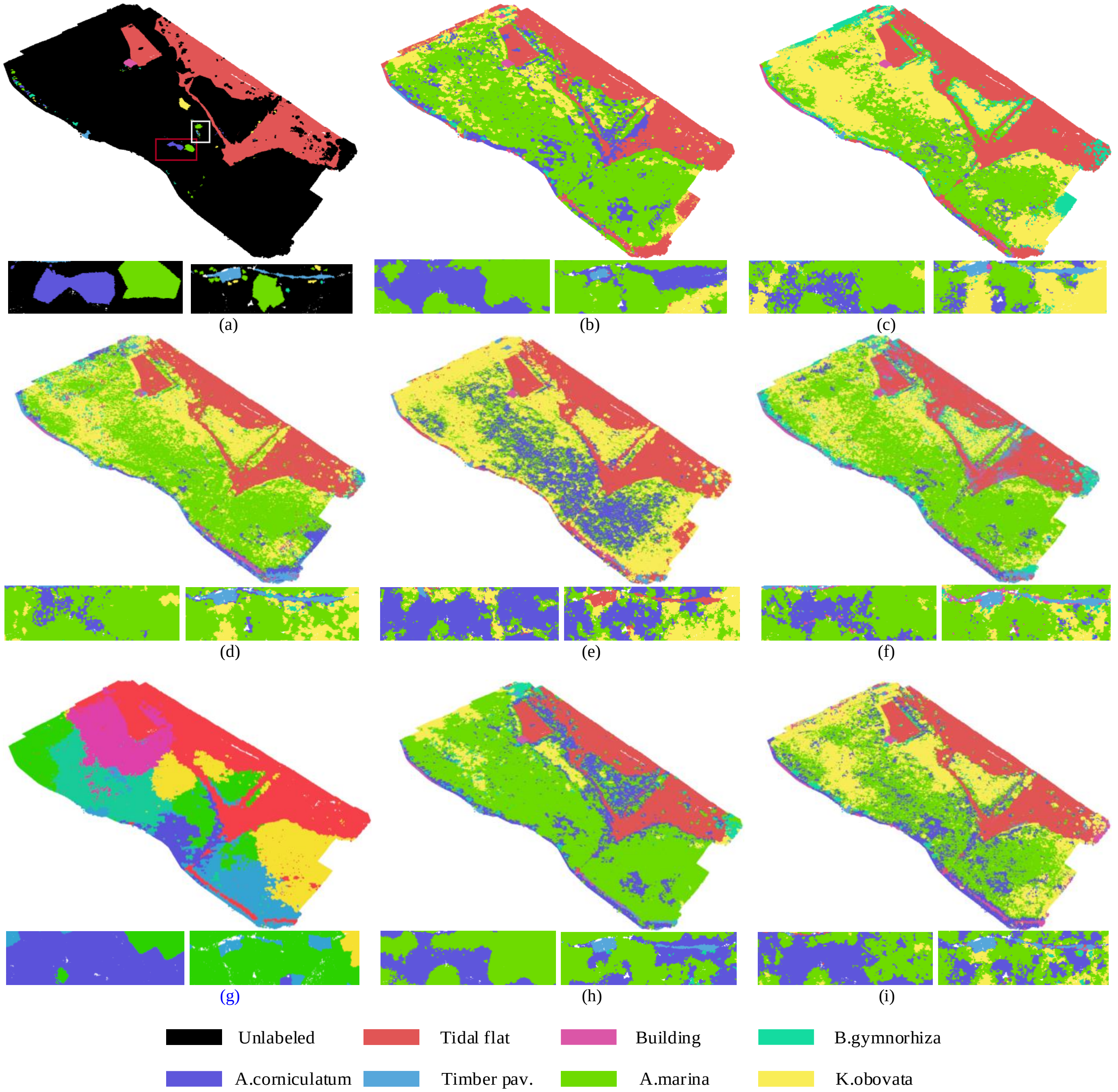

**Fig. 4.** Classification maps on the ZJKM dataset obtained by (a) Ground truth, (b) RandLA-Net, (c) CGANet, (d) RFFS-Net, (e) EyeNet, (f) MAMPC, (g) PCMamba, (h) EMPC, (i) Ours.

## C. Ablation Study

*1) Effectiveness of the Proposed Modules:* To verify the effectiveness of each module in the proposed method, we use RandLA-Net [36] as a backbone and test the effectiveness of the position-encoded global spectral feature extractor (+PGSF), the spectral-guided geometric feature extractor (+SGGF), the integrated geometric-spectral feature learning block (+IGSF), and the joint loss (Ours) on ZJKM. The comparison results are summarized in Table III. The results in Table III show that all the classification results have indeed improved when involving each module. In particular, the proposed PGSF and joint loss contribute the most significant performance improvements. Compared with the backbone [36], the complete framework yields prominent performance gains, with AA increased by 11.98% and mIoU increased by 10.53%. This validates the effectiveness of all the proposed modules.

*2) Effectiveness of the Joint Loss:* We first compare the proposed joint loss with the related center loss [45] and the common cross-entropy loss. The results in Table IV demonstrate a clear advantage of the proposed method with the joint loss over the traditional center loss and cross-entropy for both datasets. This is because the proposed joint loss incorporates a long-tail center loss and a minimum margin loss to enforce intra-class compactness and inter-class separability.

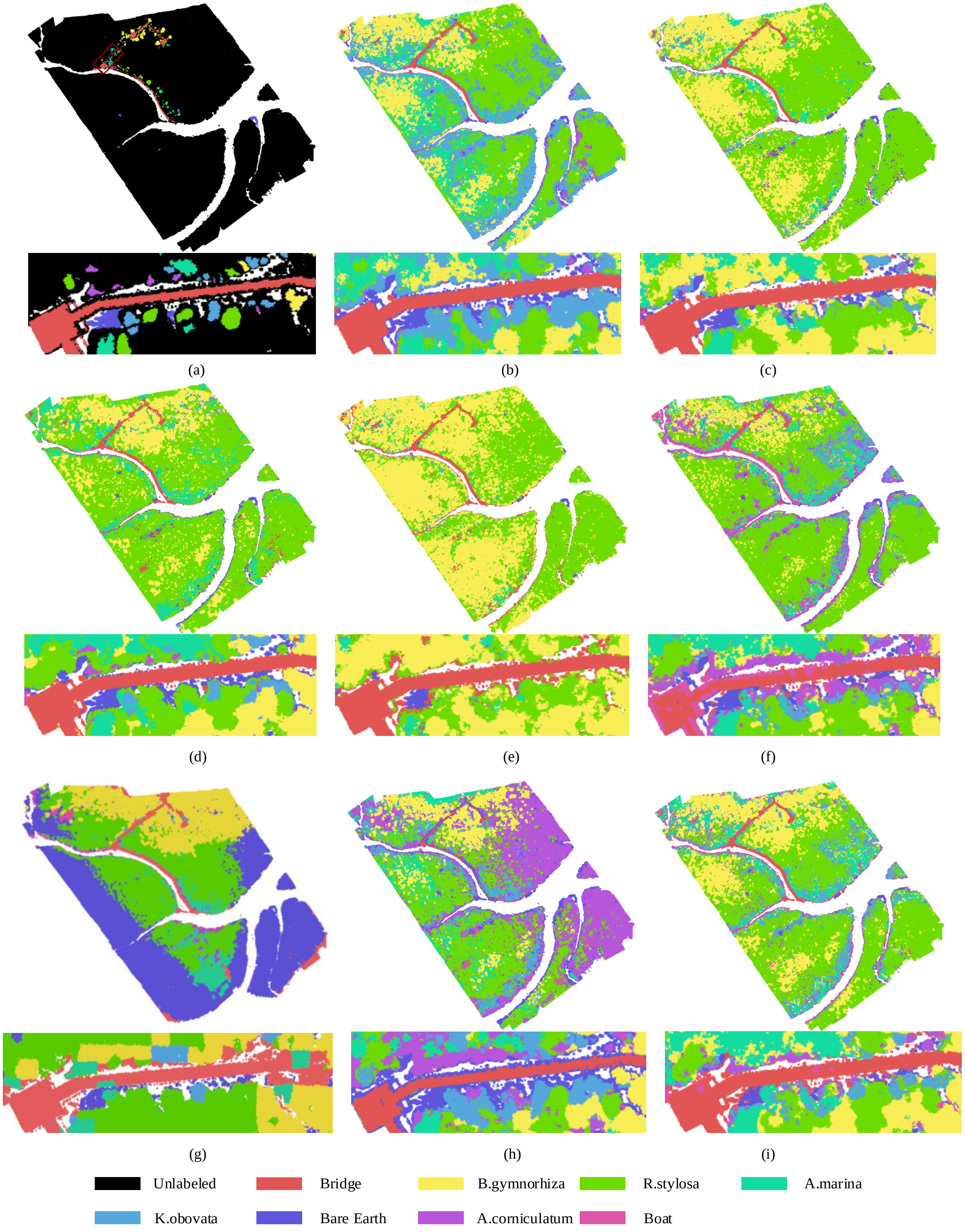

**Fig. 5.** Classification maps on the SKM dataset obtained by (a) Ground truth, (b) RandLA-Net, (c) CGANet, (d) RFFS-Net, (e) EyeNet, (f) MAMPC, (g) PCMamba, (h) EMPC, (i) Ours.

TABLE III
ABLATION RESULTS FOR DIFFERENT MODULES ON ZJKM. (1) BACKBONE, (2) ADDING THE POSITION-ENCODED GLOBAL SPECTRAL FEATURE EXTRACTOR (+PGSF), (3) ADDING SPECTRAL-GUIDED GEOMETRIC FEATURE EXTRACTOR (+SGGF), (4) ADDING INTEGRATED GEOMETRIC-SPECTRAL FEATURE LEARNING (+IGSF), (5) ADDING THE JOINT LOSS (OURS).

| Model | Backbone | +PGSF | +SGGF | +IGSF | **Ours** |
|---|---|---|---|---|---|
| PGSF | | √ | √ | √ | √ |
| SGGF | | | √ | √ | √ |
| IGSF | | | | √ | √ |
| Joint loss | | | | | √ |
| OA | 83.59 | 84.55 | 85.31 | 85.36 | 87.82 |
| **ΔOA** | **+0.00** | **+0.96** | **+1.72** | **+1.77** | **+4.23** |
| AA | 71.06 | 78.49 | 79.87 | 78.96 | 83.04 |
| **ΔAA** | **+0.00** | **+7.43** | **+8.81** | **+7.90** | **+11.98** |
| kappa | 77.79 | 79.16 | 80.15 | 80.20 | 83.60 |
| **Δkappa** | **+0.00** | **+1.37** | **+2.36** | **+2.41** | **+5.81** |
| mIoU | 60.62 | 65.38 | 66.82 | 67.14 | 71.15 |
| **ΔmIoU** | **+0.00** | **+4.76** | **+6.20** | **+6.52** | **+10.53** |

TABLE IV
COMPARISON OF THE PROPOSED JOINT LOSS AND THE RELATED CENTER LOSS ON THE ZJKM AND SKM DATASETS

| Datasets | Loss | OA | AA | kappa | mIoU |
|---|---|---|---|---|---|
| ZJKM | Cross-entropy | 85.36 | 78.96 | 80.20 | 67.14 |
| | Center | 86.95 | 81.10 | 82.37 | 69.57 |
| | Our | **87.82** | **83.04** | **83.60** | **71.15** |
| SKM | Cross-entropy | 87.02 | 77.99 | 81.80 | 64.22 |
| | Center loss | 86.94 | 77.31 | 81.63 | 63.51 |
| | Our | **87.38** | **78.52** | **82.22** | **64.90** |

TABLE V
COMPARISON OF THE COMPUTATIONAL COMPLEXITY AND TRAINING EFFICIENCY OF DIFFERENT METHODS ON THE ZJKM DATASET

| Method | Training Time (s) | #Params (×M) | FLOPs ($\times10^6$) |
|---|---|---|---|
| RandLA-Net [36] | 1561.4 | 4.99 | **1.75** |
| CGA-Net [48] | 2080.8 | 5.01 | 5.73 |
| RFFS-Net [30] | 10665.6 | 8.10 | 88.72 |
| EyeNet [49] | 4827.8 | 20.72 | 175.04 |
| MAMPC [25] | 793.4 | 4.08 | 6.62 |
| PCMamba [40] | 6493.8 | 23.45 | 95.12 |
| PTv3 [33] | 35963.0 | 5.20 | 112.48 |
| EMPC [24] | 2433.4 | **1.34** | 11.14 |
| LitePT-S [34] | 2613.0 | 12.75 | 88.38 |
| Ours | **475.5** | 5.35 | 14.86 |

We further analyze the influence of the regularization parameter on the performance of the proposed joint loss. Fig. 6 shows the joint loss versus different $\lambda \in \{10, 3, 1, 0.3, 0.01\}$ on SKM. In general, the classification performance initially increases and then declines slightly as $\lambda$ decreases. The essential reason is that larger $\lambda$ overfits the joint loss on minority and similar classes and leads to losing useful features on majority classes, while smaller $\lambda$ underfits the degree of the intra-class compactness and inter-class separability, approaching the cross-entropy loss.

*3) Analysis of the computational Efficiency:* A comparative analysis of the training time, model parameters, and FLOPs for different methods is summarized in Table V. All the experiments are conducted on a system equipped with an Intel

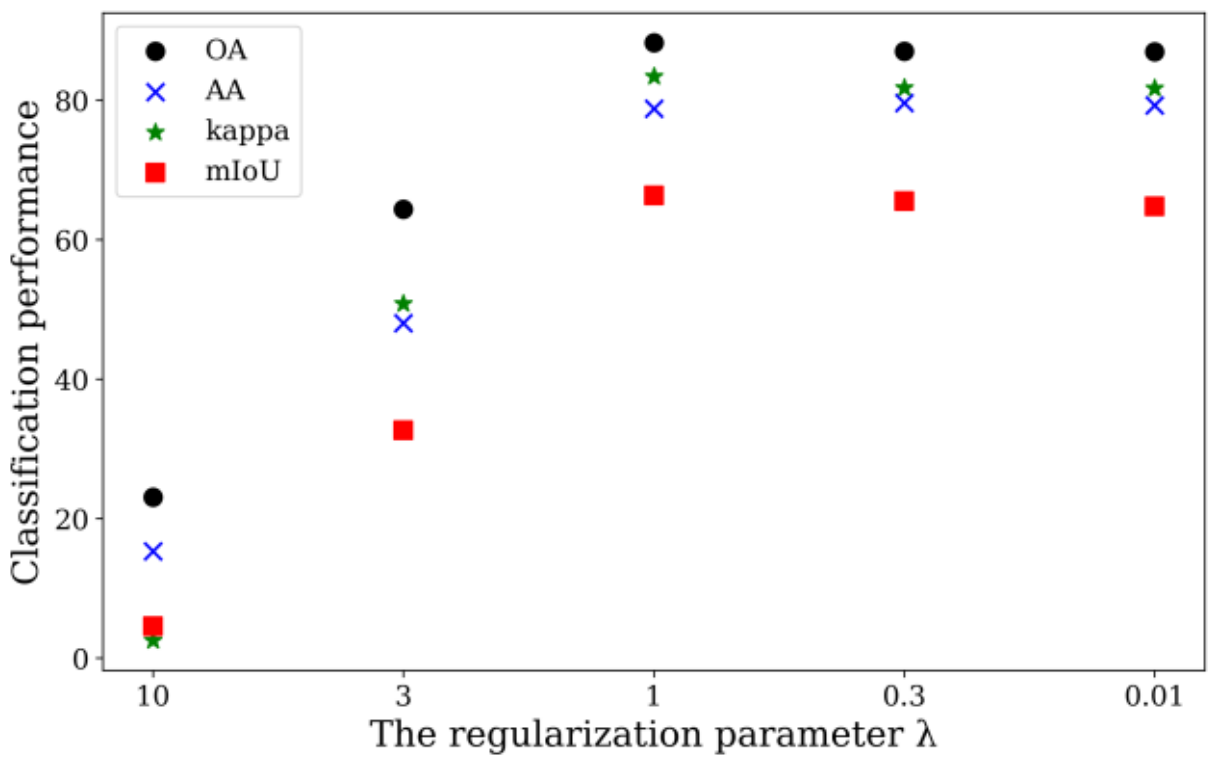

**Fig. 6.** Influence of the regularization parameter for the joint loss on SKM.

i7-12700 CPU and an RTX 3050 GPU under a Windows 10 environment. Compared to all nine reference methods (RandLA-Net [36], CGA-Net [48], RFFS-Net [30], EyeNet [49], MAMPC [25], PCMamba [40], PTv3 [33], EMPC [24], and LitePT-S [34]), the proposed method yields considerably faster training time (especially relative to RFFS-Net and PTv3) because it consists of a fully point convolution network combined with an early stopping strategy. EMPC involves much fewer parameters than our method, which is in this respect comparable to RandLA-Net, CGA-Net, MAMPC, and PTv3, and much better than EyeNet, PCMamba, and LitePT-S. Regarding FLOPs, our method is higher than RandLA-Net, CGA-Net, MAMPC, and EMPC, but considerably lower than RFFS-Net, EyeNet, PCMamba, LitePT-S, and PTv3. Given that the proposed method achieves the best classification accuracy among all compared methods (Tables I–II), it offers a favorable trade-off between computational efficiency and classification performance.

## IV. CONCLUSION

In this article, we built two airborne MPC datasets and proposed an enhanced geometric-spectral feature fusion framework based on attention for airborne multispectral point cloud classification. The proposed method simultaneously extracted position-encoded global spectral features and spectral-guided geometric features, which were fused into the integrated geometric-spectral features. In addition, we formulated an appropriate joint loss that consists of a long-tail center loss and a minimum margin loss to optimize our framework from scratch, enforcing intra-class compactness and inter-class separability. Experimental results on two self-constructed airborne datasets demonstrated state-of-the-art classification performance.

In future work, we plan to extend the proposed method along three key aspects: 1) Spectral-variability-aware augmentation for handling more complex scenes with sensor noise and illumination variations. 2) Cross-scene domain adaptation for improving the generalization across larger-scale datasets without extensive manual annotation. 3) Integration of complementary modalities (e.g., LiDAR point clouds and multispectral imagery) to enhance robustness under multimodal heterogeneous remote sensing settings.

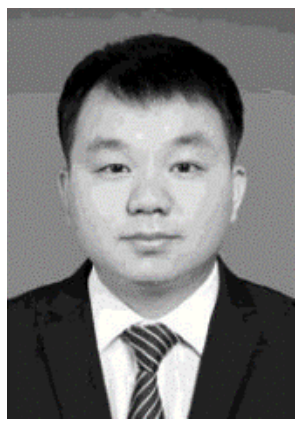

**Xian Li** (Member, IEEE) received the Ph.D. degree in instrument science and technology from Harbin Institute of Technology (HIT), China, in 2021. From 2018 to 2020, he was a Visiting Doctoral Researcher at Ghent University, Belgium, where he conducted research in the Department of Telecommunications and Information Processing with the support of the China Scholarship Council.

He is currently an Associate Professor in the School of Electronics and Information Engineering at HIT. In 2025, he was named an IEEE TGRS Best Reviewer (one of five recipients worldwide) and received the Young Elite Scientists Sponsorship Program Award from the Chinese Institute of Electronics. His research focuses on deep learning, multisource remote sensing.

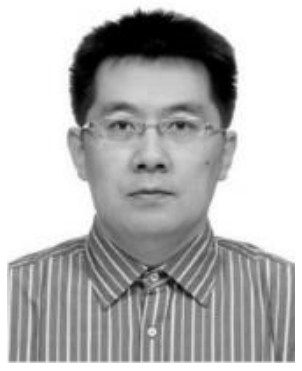

**Yanfeng Gu** (M’06-SM’16) received the Ph.D. degree in information and communication engineering from Harbin Institute of Technology, Harbin, China, in 2005. He joined as a Lecture with the School of Electronics and Information Engineering, Harbin Institute of Technology (HIT). He was appointed as Associate Professor at the same institute in 2006; meanwhile, he was enrolled in first Outstanding Young Teacher Training Program of HIT. From 2011 to 2012, he was a Visiting Scholar with the Department of Electrical Engineering and Computer Science, University of California, Berkeley, CA, USA. He is currently a Professor with the Department of Information Engineering, HIT, Harbin, China. He has published more than 100 peer-reviewed papers, four book chapters, and he is the inventor or coinventor of 20 patents. His research interests include space intelligent remote sensing and information processing, multimodal hyperspectral remote sensing, spaceborne time-series image processing.

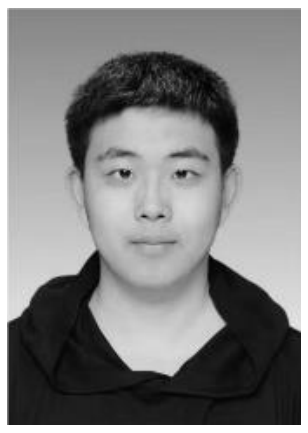

**Linghua Xu** is currently pursuing the B.S. degree in remote sensing science and technology at Harbin Institute of Technology, Harbin, China. He has authored or coauthored one EI-indexed paper and has two patent applications under examination. He has been awarded the Heilongjiang Provincial Natural Science Foundation Undergraduate Project. His current research interests include full-waveform LiDAR data processing, remote sensing data analysis, and intelligent remote sensing information processing.

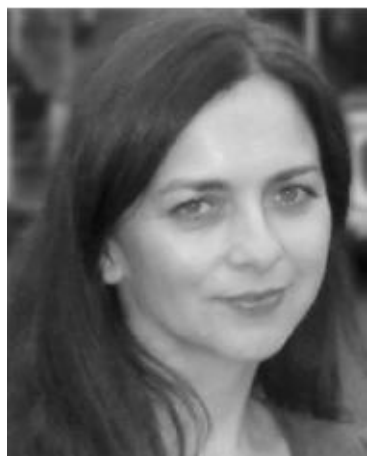

**Aleksandra Pižurica** (Senior Member, IEEE) received the Diploma degree in electrical engineering from the University of Novi Sad, Novi Sad, Serbia, in 1994, the M.Sc. degree in telecommunications from the University of Belgrade, Belgrade, Serbia, in 1997, and the Ph.D. degree in engineering from Ghent University, Ghent, Belgium, in 2002.

She is currently a Professor of statistical image modeling with Ghent University. Her research interests include the area of signal and image processing and machine learning, including multiresolution statistical image models, Markov random field models, sparse coding, representation learning, and image and video reconstruction, restoration, and analysis. Prof. Pižurica received the scientific prize “de Boelpaepe” for 2013–2014, awarded by the Royal Academy of Science, Letters and Fine Arts of Belgium for her contributions to statistical image modeling and applications to digital painting analysis. The work of her team has been awarded twice the Best Paper Award of the IEEE Geoscience and Remote Sensing Society Data Fusion Contest in 2013 and 2014. She has served as an Associate Editor for the IEEE TRANSACTIONS ON IMAGE PROCESSING from 2012 to 2016 and a Senior Area Editor for the IEEE TRANSACTIONS ON IMAGE PROCESSING from 2016 to 2019. She was a Lead Guest Editor of the EURASIP Journal on Advances in Signal Processing for the Special Issue “Advanced Statistical Tools for Enhanced Quality Digital Imaging with Realistic Capture Models” in 2013. She is also an Associate Editor of the IEEE TRANSACTIONS ON CIRCUITS AND SYSTEMS FOR VIDEO TECHNOLOGY.